\documentclass[letterpaper]{article} 
\usepackage{aaai2027}  

\makeatletter
\gdef\copyright@on{}
\gdef\copyright@text{}
\makeatother
\usepackage[hyphens]{url}  
\usepackage{graphicx} 
\usepackage{natbib}  
\usepackage{caption} 
\usepackage{algorithm}
\usepackage{algorithmic}

\usepackage{tikz}
\usetikzlibrary{arrows.meta, positioning, calc, fit, backgrounds}
\usepackage{newfloat}
\usepackage{listings}
\DeclareCaptionStyle{ruled}{labelfont=normalfont,labelsep=colon,strut=off} 
\floatstyle{ruled}
\newfloat{listing}{tb}{lst}{}
\floatname{listing}{Listing}

\usepackage{booktabs}

\usepackage{amsmath}
\usepackage{amssymb}
\usepackage{mathtools}
\usepackage{amsthm}
\usepackage{multirow}
\usepackage{adjustbox}
\title{NPMixer: Multi-Scale Neighboring Patch Mixing for Time Series Forecasting}

\author{
    Jung Min Choi\thanks{Corresponding author: \texttt{choi@ismll.de}},
    Ibram Abdelmalak,
    Ngoc Son Le,
    Vijaya Krishna Yalavarthi,
    Lars Schmidt-Thieme
}

\affiliations{
    ISMLL, University of Hildesheim, VWFS Data Analytics Research Center
    (VWFS-DARC), Hildesheim, Germany\\
    \texttt{choi@ismll.de},
    \texttt{abdelmalak@ismll.de},
    \texttt{sle@ismll.de},
    \texttt{yalavarthi@ismll.de},
    \texttt{schmidt-thieme@ismll.de}
}

\begin{document}

\makeatletter
\gdef\copyright@on{}
\gdef\copyright@text{}
\makeatother
\maketitle

\begin{abstract}
    Multivariate time series forecasting requires models to capture patterns at different temporal scales, connect local observations with broader temporal structure, and exploit cross-variable dependencies. Existing patch-based approaches often enlarge their temporal context through overlapping windows, which repeatedly encode the same observations but do not explicitly organize the transition from local patterns to long-range dependencies. Moreover, using one fixed form of channel interaction for every dataset overlooks that predictive cross-variable relationships may be sparse, dense, or even unnecessary. 
    These considerations motivate NPMixer, a Multi-Scale \textbf{N}eighboring \textbf{P}atch \textbf{Mixer}. The Hierarchical Neighboring Mixer Block progressively aggregates adjacent non-overlapping patches, establishing an explicit local-to-global modeling path while avoiding the redundancy of dense sliding windows. To accommodate dataset-dependent channel relationships, NPMixer supports selective Top-$k$ interaction as well as broader MLP-based mixing instead of enforcing one channel-dependency assumption. It further employs the Stationary Wavelet Transform to represent temporal patterns at multiple aligned resolutions. Together, these components provide a unified framework for modeling local, global, multi-scale, and cross-variable structures. Averaged across forecasting horizons, NPMixer ranks first in MSE on six of the seven long-term benchmarks; for MAE, it attains or shares the top rank on all seven. On the four PEMS benchmarks, it also produces more MSE-leading configurations than any competing method. The ablation results show that multi-scale decomposition, local patch modeling, hierarchical neighboring aggregation, and configurable channel interaction each contribute to the final performance.
    \end{abstract}

\section{Introduction}

\begin{figure}[t]
    \centering
    \includegraphics[width=0.85\columnwidth]{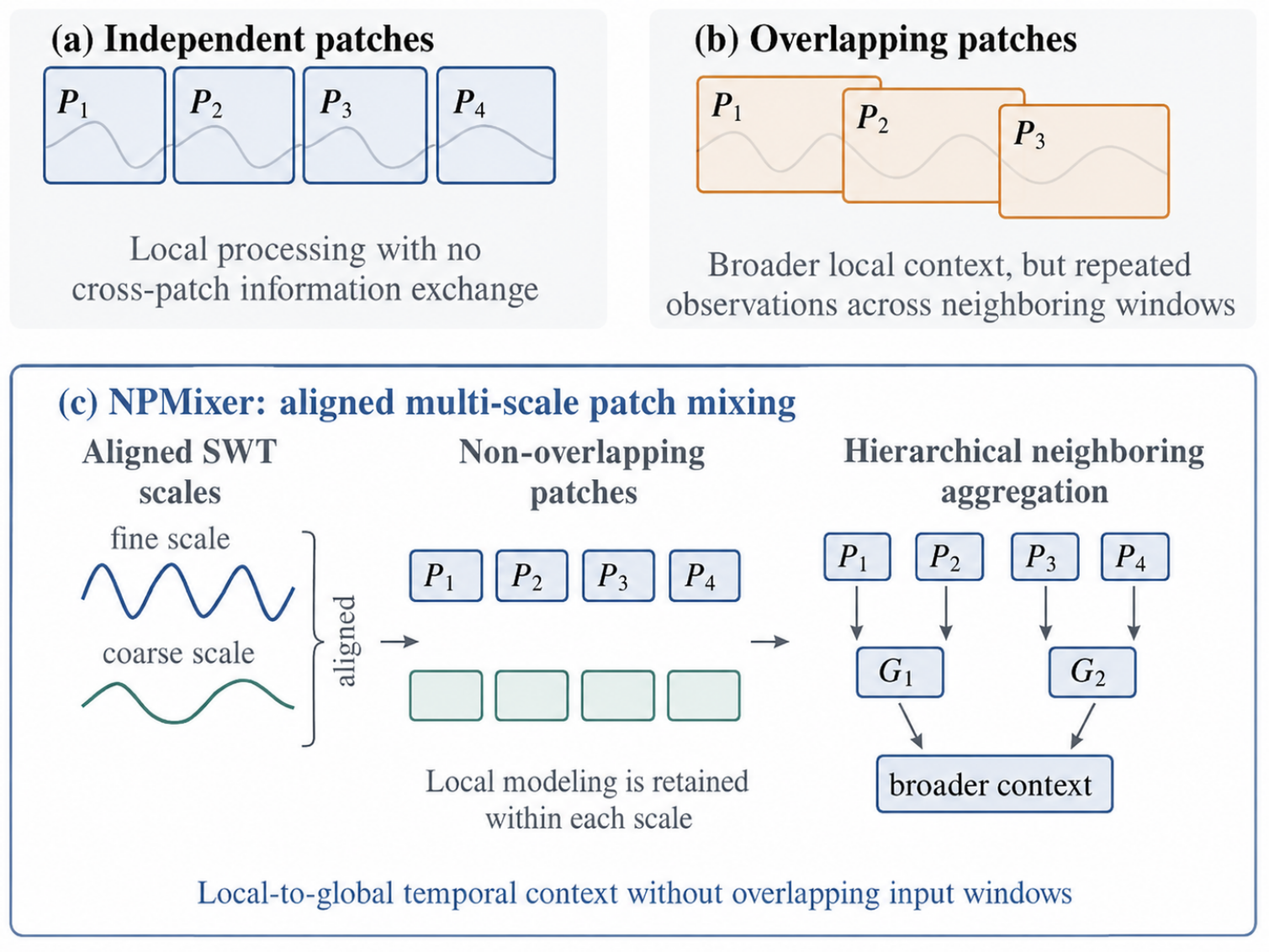}
    \caption{
Motivation for hierarchical neighboring-patch aggregation.
Non-overlapping patches avoid redundant encoding but limit cross-patch
interaction. NPMixer progressively aggregates adjacent patch groups across
aligned SWT scales to capture broader temporal context.
}
    \label{fig:npmixer_motivation}
\end{figure}

Applications of time series forecasting (TSF) range from climate modeling \cite{weather} and energy-demand prediction \cite{electricity} to transportation monitoring \cite{pems}.
Deep neural forecasters have increasingly emerged as alternatives to traditional statistical techniques such as ARIMA \cite{arima}.
Transformer architectures \cite{attention}, including Informer \cite{informer}, FEDformer \cite{fedformer}, and Crossformer \cite{crossformer}, model long-range temporal patterns and dependencies among variables.
In parallel, MLP-based methods such as DLinear \cite{dlinear}, TSMixer \cite{tsmixer}, and TimeMixer \cite{timemixer} show that comparatively lightweight architectures can attain competitive forecasting accuracy.

Patching, adapted from Vision Transformers~\cite{vit}, has become
an effective representation strategy for time series forecasting.
Instead of treating individual observations as tokens, patch-based
models represent contiguous subsequences as local temporal units.
This can preserve local patterns, shorten the token sequence, and
permit the model to process longer historical contexts
\cite{patchtst,timebridge}. However, the choice of patch construction
also affects how temporal information is exchanged. Independently
processing non-overlapping patches limits interaction across patch
boundaries, whereas overlapping patches improve local continuity by
repeatedly encoding some of the same observations, thereby increasing
the number of tokens.

Figure~\ref{fig:npmixer_motivation} illustrates how NPMixer addresses
this trade-off. Rather than introducing overlapping sliding windows,
NPMixer uses non-overlapping patches and progressively mixes adjacent
patch groups. This hierarchical aggregation expands the receptive
field from individual patches to increasingly broader temporal
contexts while retaining a compact patch representation.

Frequency-domain methods further expose temporal structure
through the Fast Fourier Transform \cite{frets,msgnet} or
wavelet decomposition \cite{waveform,simpletm}, separating
patterns associated with different periodicities and resolutions
\cite{fedformer}. NPMixer obtains its multi-resolution representation with the
Stationary Wavelet Transform (SWT). Unlike the Discrete Wavelet
Transform \cite{dwt}, SWT performs no downsampling and therefore
keeps the original sequence length at every resolution \cite{swt}.
This temporal alignment allows approximation and detail
coefficients to share the same patch-processing pipeline.

We introduce the \textbf{Multi-Scale Neighboring Patch Mixer}
(\textbf{NPMixer}). At each SWT scale, NPMixer applies
non-overlapping patching, an intra-patch MLP, a channel mixer,
and a Hierarchical Neighboring Mixer. The hierarchy recursively
exchanges information between adjacent blocks of increasing size
through gated relational updates. For cross-variate interaction,
we study a dense MLP mixer and a sparse online Top-$k$ mixer
that aggregates only the most relevant channels. The processed
scales are fused directly before prediction, without inverse SWT.

Our main contributions are:
\begin{enumerate}
    \item We propose a hierarchical non-overlapping patch mixer
    that connects local patch semantics to broader temporal
    dependencies without redundant sliding windows.

    \item We introduce an SWT-based multi-scale forecasting
    framework with temporally aligned scale processing, learnable
    fusion, and controllable dense or sparse channel interaction.

    \item We demonstrate that NPMixer achieves state-of-the-art
    or highly competitive performance across long-term and
    short-term multivariate forecasting benchmarks.
\end{enumerate}
\section{Related Work}

\textbf{Time Series Forecasting.}
Modern TSF models are commonly based on MLPs, Transformers,
CNNs, or GNNs. Simple MLP architectures such as
DLinear~\cite{dlinear} and TSMixer~\cite{tsmixer} achieve
competitive forecasting performance at low computational cost,
while TimeMixer~\cite{timemixer} combines MLP-based modeling
with multi-scale temporal decomposition. Transformer models
capture long-range dependencies through attention. Informer
improves attention efficiency for long sequences~\cite{informer},
Autoformer introduces decomposition and auto-correlation
mechanisms~\cite{autoformer}, and FEDformer incorporates
frequency-domain representations~\cite{fedformer}. PatchTST
represents temporal segments as tokens~\cite{patchtst}, whereas
iTransformer applies attention over variate tokens to model
cross-channel dependencies~\cite{itransformer}. CNN-based
models, including MICN~\cite{micn} and TimesNet~\cite{timesnet},
capture local and multi-period patterns. GNN-based approaches
such as Graph WaveNet~\cite{graphwavenet},
MTGNN~\cite{mtgnn}, MSGNet~\cite{msgnet}, and
WaveForM~\cite{waveform} model variables as graph nodes and
learn temporal or cross-variate relations through graph
propagation.

\textbf{Patching in TSF.}
Inspired by the Vision Transformer~\cite{vit}, patching
represents a time series as sub-sequence tokens, preserving local
patterns while reducing the effective sequence length.
PatchTST~\cite{patchtst} applies patch tokens within a
Transformer, and Crossformer~\cite{crossformer} uses
dimension-segment-wise embeddings to model cross-time and
cross-variable dependencies. MLP-based approaches also exploit
patch representations: TSMixer~\cite{tsmixer2} mixes information
within and across patches, while HDMixer~\cite{hdmixer} uses
extendable patches and hierarchical dependency modeling.

Many patch-based forecasters use overlapping windows to preserve
local continuity, but this repeatedly encodes the same
observations and increases computational cost. The Swin
Transformer~\cite{swintransformer} demonstrates that
non-overlapping windows can be combined with hierarchical
aggregation. Following this principle, NPMixer uses
non-overlapping temporal patches and progressively aggregates
neighboring blocks to expand the receptive field without dense
sliding windows.

\textbf{Frequency-Domain TSF.}
Transforming a sequence into the spectral domain can make
periodic behavior and long-range structure easier to identify
than in its original temporal representation. FEDformer~\cite{fedformer} integrates
frequency components into a decomposed Transformer, FiLM
preserves historical information through frequency-enhanced
memory modeling~\cite{film}, and FreTS applies MLPs directly in
the frequency domain~\cite{frets}. StemGNN~\cite{stemgnn}
combines graph and temporal spectral modeling to capture
inter-series and temporal dependencies.

Wavelet methods instead represent signals at multiple temporal
resolutions. WaveForM~\cite{waveform} combines wavelet
decomposition with graph learning, but the downsampling in the
standard Discrete Wavelet Transform can disrupt temporal
alignment and shift invariance. SimpleTM~\cite{simpletm}
addresses this issue using a learnable Stationary Wavelet
Transform (SWT), which preserves the original sequence length at
every scale. NPMixer likewise uses SWT to obtain aligned
approximation and detail coefficients, but processes them using
scale-wise neighboring patch mixing.
\section{Methodology}

NPMixer captures multi-scale temporal dynamics and cross-channel
dependencies for long-term time series forecasting. Its overall
architecture is shown in Figure~\ref{fig:npmixer_main_architecture}.

After RevIN normalization~\cite{revin}, a fixed
\textbf{Stationary Wavelet Transform (SWT)} decomposes the input into
temporally aligned approximation and detail coefficients. NPMixer omits
inverse SWT and processes each coefficient independently using the same
scale-wise architecture.

At each scale, the coefficient is divided into non-overlapping patches.
An \textbf{Intra-Patch MLP} models local temporal patterns, followed by
either the sparse \textbf{Top-$k$ Channel Mixer} or the dense
\textbf{MLP Channel Mixer} for cross-variate interaction. The
\textbf{Hierarchical Neighboring Mixer Block} then progressively
aggregates adjacent patches to expand the receptive field from local
segments to broader temporal contexts.

The scale-wise representations are projected back to the look-back
length and combined through \textbf{Learnable Scale Fusion}. A residual
sequence-level MLP and final prediction layer generate the forecast,
which is restored to the original scale through RevIN denormalization.

\subsection{Preliminaries}

Consider a multivariate time series with $C$ variables observed
over a historical look-back window of length $L$. The input is
represented as $\mathbf{X} \in \mathbb{R}^{C \times L}$, where
$\mathbf{x}_t =
[x_{t,1},x_{t,2},\ldots,x_{t,C}]^\top \in \mathbb{R}^{C}$
contains the observations of all variables at time step $t$.

Given $\mathbf{X}$, the forecasting task is to predict the next
$H$ time steps, denoted by
$\mathbf{Y} \in \mathbb{R}^{C \times H}$. NPMixer learns a
mapping
\[
    \hat{\mathbf{Y}}
    =
    \mathcal{F}(\mathbf{X}),
    \qquad
    \mathcal{F}:
    \mathbb{R}^{C \times L}
    \rightarrow
    \mathbb{R}^{C \times H},
\]
such that the prediction $\hat{\mathbf{Y}}$ approximates the
ground-truth future sequence $\mathbf{Y}$.

\begin{figure*}[t]
    \centering
    \resizebox{\textwidth}{!}{%
    \begin{tikzpicture}[
        >=Stealth,
        node distance=1.0cm and 0.7cm,
        font=\sffamily\small,
        op_pink/.style={
            draw=red!80!black, fill=red!10, thick, rounded corners=5pt,
            minimum height=1cm, minimum width=1.9cm, align=center
        },
        op_orange/.style={
            draw=orange!80!black, fill=orange!10, thick, rounded corners=5pt,
            minimum height=1cm, minimum width=2.15cm, align=center
        },
        op_green/.style={
            draw=green!60!black, fill=green!10, thick, rectangle, rounded corners=2pt,
            minimum height=1cm, minimum width=2.25cm, align=center
        },
        op_blue/.style={
            draw=cyan!60!black, fill=cyan!10, thick, rectangle, rounded corners=2pt,
            minimum height=1cm, minimum width=1.7cm, align=center
        },
        op_violet/.style={
            draw=violet!80!black, fill=violet!10, thick, rounded corners=4pt,
            minimum height=1cm, minimum width=2.55cm, align=center
        },
        scale_node/.style={
            draw=gray!70!black, fill=gray!10, thick, rounded corners=3pt,
            minimum height=0.8cm, minimum width=1.55cm, align=center
        },
        plain/.style={
            draw=none, fill=none, font=\bfseries
        },
        edge_label/.style={
            font=\scriptsize\itshape, fill=white, inner sep=1pt, text=gray!30!black
        }
    ]
  
        \node[plain] (Input) {Input $X_t$};
        \node[op_pink, right=0.8cm of Input] (RevIn) {RevIN.norm};
        \node[op_orange, right=0.8cm of RevIn] (SWT) {SWT\\Decomposition};
  
        \draw[->, thick] (Input) -- (RevIn);
        \draw[->, thick] (RevIn) -- (SWT);
  
        \coordinate (BranchStart) at ($(SWT.east) + (1.6, 0)$);
  
        \node[scale_node] (ScaleA) at ($(BranchStart) + (0, 2.15)$) {$A_J$};
        \node[op_blue, right=0.6cm of ScaleA] (PatchA) {Patching};
        \node[op_green, right=0.6cm of PatchA] (IntraA) {Intra-Patch\\MLP};
        \node[op_violet, right=0.6cm of IntraA] (ChanA) {Channel\\Mixer};
        \node[op_green, right=0.6cm of ChanA] (HierA) {Hierarchical\\Mixer Block};
        \node[op_orange, right=0.6cm of HierA] (ProjA) {Projection};
  
        \node[scale_node] (ScaleM) at (BranchStart) {$D_m$};
        \node[op_blue, right=0.6cm of ScaleM] (PatchM) {Patching};
        \node[op_green, right=0.6cm of PatchM] (IntraM) {Intra-Patch\\MLP};
        \node[op_violet, right=0.6cm of IntraM] (ChanM) {Channel\\Mixer};
        \node[op_green, right=0.6cm of ChanM] (HierM) {Hierarchical\\Mixer Block};
        \node[op_orange, right=0.6cm of HierM] (ProjM) {Projection};
  
        \node[scale_node] (ScaleD) at ($(BranchStart) + (0, -2.15)$) {$D_1$};
        \node[op_blue, right=0.6cm of ScaleD] (PatchD) {Patching};
        \node[op_green, right=0.6cm of PatchD] (IntraD) {Intra-Patch\\MLP};
        \node[op_violet, right=0.6cm of IntraD] (ChanD) {Channel\\Mixer};
        \node[op_green, right=0.6cm of ChanD] (HierD) {Hierarchical\\Mixer Block};
        \node[op_orange, right=0.6cm of HierD] (ProjD) {Projection};
  
        \node at ($(ScaleA)!0.5!(ScaleM)$) {$\vdots$};
        \node at ($(ScaleM)!0.5!(ScaleD)$) {$\vdots$};
  
        \coordinate (Split) at ($(SWT.east) + (0.55, 0)$);
        \draw[thick] (SWT.east) -- (Split);
  
        \draw[->, thick] (Split) |- node[edge_label, pos=0.72] {$C_0=A_J$} (ScaleA.west);
        \draw[->, thick] (Split) -- node[edge_label] {$C_m$} (ScaleM.west);
        \draw[->, thick] (Split) |- node[edge_label, pos=0.72] {$C_J=D_1$} (ScaleD.west);
  
        \draw[->, thick] (ScaleA) -- (PatchA);
        \draw[->, thick] (PatchA) -- (IntraA);
        \draw[->, thick] (IntraA) -- (ChanA);
        \draw[->, thick] (ChanA) -- (HierA);
        \draw[->, thick] (HierA) -- (ProjA);
  
        \draw[->, thick] (ScaleM) -- (PatchM);
        \draw[->, thick] (PatchM) -- (IntraM);
        \draw[->, thick] (IntraM) -- (ChanM);
        \draw[->, thick] (ChanM) -- (HierM);
        \draw[->, thick] (HierM) -- (ProjM);
  
        \draw[->, thick] (ScaleD) -- (PatchD);
        \draw[->, thick] (PatchD) -- (IntraD);
        \draw[->, thick] (IntraD) -- (ChanD);
        \draw[->, thick] (ChanD) -- (HierD);
        \draw[->, thick] (HierD) -- (ProjD);
  
        \node[op_violet, right=1.25cm of ProjM, minimum width=2.8cm] (ScaleFusion)
        {Learnable\\Scale Fusion\\$\sum_m \alpha_m \tilde{C}_m$};
  
        \node[edge_label, above=0.12cm of ScaleFusion] 
        {$\alpha_m=\mathrm{softmax}(\ell_m)$};
  
        \draw[->, thick] (ProjA.east) -| ++(0.45,0) |- (ScaleFusion.west);
        \draw[->, thick] (ProjM.east) -- (ScaleFusion.west);
        \draw[->, thick] (ProjD.east) -| ++(0.45,0) |- (ScaleFusion.west);
  
        \node[op_green, right=0.85cm of ScaleFusion] (ResidualMLP) {Sequence MLP\\+ Residual};
        \node[op_pink, right=0.85cm of ResidualMLP] (Pred) {Prediction\\Layer};
        \node[op_pink, right=0.7cm of Pred] (RevDe) {RevIN.denorm};
        \node[plain, right=0.7cm of RevDe] (Out) {Output $X_T$};
  
        \draw[->, thick] (ScaleFusion) -- (ResidualMLP);
        \draw[->, thick] (ResidualMLP) -- (Pred);
        \draw[->, thick] (Pred) -- (RevDe);
        \draw[->, thick] (RevDe) -- (Out);
  
        \begin{scope}[on background layer]
            \node[
                fit=(ScaleA)(PatchA)(IntraA)(ChanA)(HierA)(ProjA)
                    (ScaleD)(PatchD)(IntraD)(ChanD)(HierD)(ProjD),
                fill=gray!5, draw=gray!50, dashed, thick, rounded corners,
                inner xsep=13pt,
                inner ysep=24pt,
                label={
                    [anchor=south, fill=white, text=violet!90!black,
                     font=\bfseries\large, yshift=3pt]
                    north:Multi-Scale Processing
                }
            ] (Background) {};
        \end{scope}
  
    \end{tikzpicture}
    }
    \caption{Overall architecture of NPMixer. RevIN-normalized inputs
are decomposed by SWT, processed at each scale using patch-based
temporal modeling and the selected channel mixer, fused across
scales, and mapped to the forecast horizon before RevIN
denormalization.}
    \label{fig:npmixer_main_architecture}
  \end{figure*}
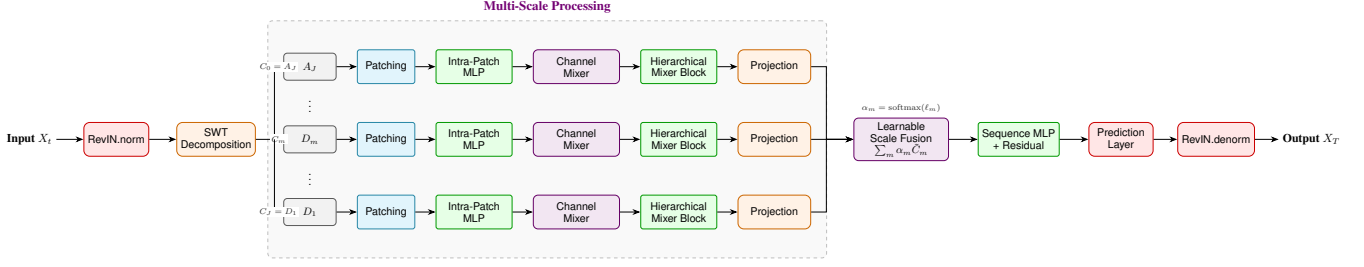
  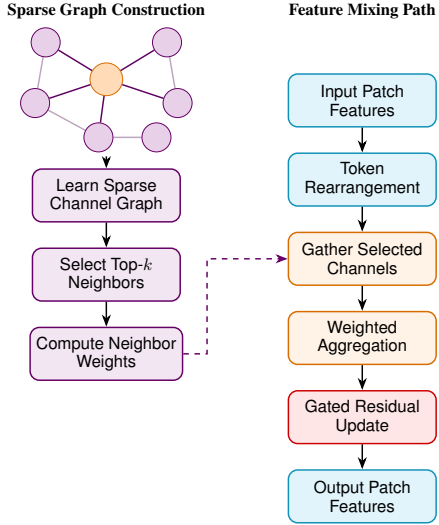
\begin{figure}[t]
    \centering
    \resizebox{0.70\columnwidth}{!}{%
    \begin{tikzpicture}[
        >=Stealth,
        font=\sffamily\small,
        box/.style={
            rounded corners=5pt,
            thick,
            minimum width=2.8cm,
            minimum height=1.0cm,
            align=center
        },
        graphbox/.style={
            box,
            draw=violet!80!black,
            fill=violet!10
        },
        featbox/.style={
            box,
            draw=cyan!70!black,
            fill=cyan!10
        },
        mixbox/.style={
            box,
            draw=orange!85!black,
            fill=orange!10
        },
        updatebox/.style={
            box,
            draw=red!80!black,
            fill=red!10
        },
        title/.style={
            draw=none,
            fill=none,
            font=\bfseries\small
        },
        gnode/.style={
            circle,
            draw=violet!80!black,
            fill=violet!18,
            minimum size=5.5mm,
            inner sep=0pt
        },
        gcenter/.style={
            circle,
            draw=orange!85!black,
            fill=orange!28,
            minimum size=6.2mm,
            inner sep=0pt
        },
        gedge/.style={
            thick,
            violet!75!black
        },
        weakedge/.style={
            thick,
            violet!50!black,
            opacity=0.35
        }
    ]

    \node[title] at (0,4.45) {Sparse Graph Construction};
    \node[title] at (4.85,4.45) {Feature Mixing Path};

    \node[gcenter] (gc) at (0,3.15) {};

    \node[gnode] (g1) at (-1.15,3.85) {};
    \node[gnode] (g2) at ( 1.15,3.85) {};
    \node[gnode] (g3) at (-1.35,2.70) {};
    \node[gnode] (g4) at ( 1.35,2.70) {};
    \node[gnode] (g5) at (-0.15,2.05) {};
    \node[gnode] (g6) at ( 0.95,2.05) {};

    \draw[gedge] (gc) -- (g1);
    \draw[gedge] (gc) -- (g2);
    \draw[gedge] (gc) -- (g3);
    \draw[gedge] (gc) -- (g4);
    \draw[gedge] (gc) -- (g5);

    \draw[weakedge] (g1) -- (g3);
    \draw[weakedge] (g2) -- (g4);
    \draw[weakedge] (g5) -- (g6);
    \draw[weakedge] (g3) -- (g5);

    \node[draw=none, fit=(gc)(g1)(g2)(g3)(g4)(g5)(g6), inner sep=2pt] (GraphIcon) {};

    \node[graphbox] (GraphLearn) at (0,0.95)
    {Learn Sparse\\Channel Graph};

    \node[graphbox] (TopK) at (0,-0.55)
    {Select Top-$k$\\Neighbors};

    \node[graphbox] (Weights) at (0,-2.05)
    {Compute Neighbor\\Weights};

    \draw[->, thick] (GraphIcon.south) -- (GraphLearn.north);
    \draw[->, thick] (GraphLearn) -- (TopK);
    \draw[->, thick] (TopK) -- (Weights);

    \node[featbox] (Input) at (4.85,2.75)
    {Input Patch\\Features};

    \node[featbox] (Rearrange) at (4.85,1.25)
    {Token\\Rearrangement};

    \node[mixbox] (Gather) at (4.85,-0.25)
    {Gather Selected\\Channels};

    \node[mixbox] (Aggregate) at (4.85,-1.75)
    {Weighted\\Aggregation};

    \node[updatebox] (Gate) at (4.85,-3.25)
    {Gated Residual\\Update};

    \node[featbox] (Output) at (4.85,-4.75)
    {Output Patch\\Features};

    \draw[->, thick] (Input) -- (Rearrange);
    \draw[->, thick] (Rearrange) -- (Gather);
    \draw[->, thick] (Gather) -- (Aggregate);
    \draw[->, thick] (Aggregate) -- (Gate);
    \draw[->, thick] (Gate) -- (Output);

    \draw[->, thick, dashed, violet!80!black]
        (Weights.east) -- ++(0.45,0) |- node[pos=0.30, above, font=\scriptsize\itshape]
        {} (Gather.west);

    \end{tikzpicture}
    }
    \caption{Architecture of the top-$k$ channel mixer in NPMixer. The module learns a sparse channel graph, selects the top-$k$ related channels, and uses the resulting indices and weights to aggregate channel features through a gated residual update.}
    \label{fig:topk_channel_mixer}
\end{figure}

  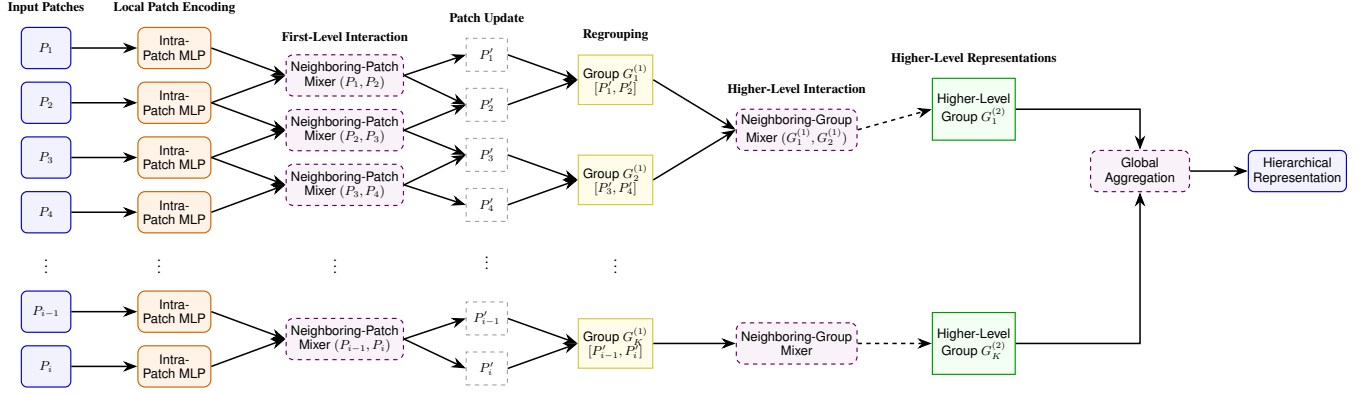
\begin{figure*}[t]
    \centering
    \resizebox{\textwidth}{!}{%
    \begin{tikzpicture}[
        node distance=1.5cm and 1.5cm,
        font=\small\sffamily,
        patch/.style={
            draw=blue!80!black,
            fill=blue!5,
            thick,
            rectangle,
            rounded corners,
            minimum height=1cm,
            minimum width=1.2cm,
            align=center
        },
        smlp/.style={
            draw=orange!80!black,
            fill=orange!10,
            thick,
            rectangle,
            rounded corners=5pt,
            minimum height=1cm,
            minimum width=1.2cm,
            align=center
        },
        mix/.style={
            draw=violet!80!black,
            fill=violet!5,
            thick,
            dashed,
            rectangle,
            rounded corners=5pt,
            minimum height=1cm,
            minimum width=2.4cm,
            align=center
        },
        block/.style={
            draw=yellow!80!black,
            fill=yellow!10,
            thick,
            rectangle,
            minimum height=1.2cm,
            minimum width=1.8cm,
            align=center
        },
        mega/.style={
            draw=green!60!black,
            fill=green!5,
            thick,
            rectangle,
            minimum height=1.5cm,
            minimum width=2.0cm,
            align=center
        },
        dissolved/.style={
            draw=gray!80,
            fill=white,
            thin,
            dashed,
            rectangle,
            minimum height=0.8cm,
            minimum width=1cm,
            align=center
        },
        dots/.style={
            draw=none,
            fill=none,
            align=center,
            font=\large
        },
        group_label/.style={
            font=\bfseries\footnotesize,
            anchor=south,
            yshift=0.2cm
        },
        flow/.style={
            -{Stealth[length=3.5mm,width=2.6mm]},
            line width=1.05pt
        },
        flow dashed/.style={
            flow,
            dashed
        }
    ]

    \node[patch] (P1) {$P_1$};
    \node[patch, below=0.3cm of P1] (P2) {$P_2$};
    \node[patch, below=0.3cm of P2] (P3) {$P_3$};
    \node[patch, below=0.3cm of P3] (P4) {$P_4$};
    \node[dots, below=0.3cm of P4] (D1) {$\vdots$};
    \node[patch, below=0.3cm of D1] (Pi_prev) {$P_{i-1}$};
    \node[patch, below=0.3cm of Pi_prev] (Pi) {$P_i$};

    \node[group_label] at (P1.north) {Input Patches};

    \node[smlp, right=1.6cm of P1] (S1)
        {Intra-\\Patch MLP};

    \node[smlp, right=1.6cm of P2] (S2)
        {Intra-\\Patch MLP};

    \node[smlp, right=1.6cm of P3] (S3)
        {Intra-\\Patch MLP};

    \node[smlp, right=1.6cm of P4] (S4)
        {Intra-\\Patch MLP};

    \node[dots, right=2.4cm of D1] (D2)
        {$\vdots$};

    \node[smlp, right=1.6cm of Pi_prev] (Si_prev)
        {Intra-\\Patch MLP};

    \node[smlp, right=1.6cm of Pi] (Si)
        {Intra-\\Patch MLP};

    \node[group_label] at (S1.north)
        {Local Patch Encoding};

    \draw[flow] (P1.east) -- (S1.west);
    \draw[flow] (P2.east) -- (S2.west);
    \draw[flow] (P3.east) -- (S3.west);
    \draw[flow] (P4.east) -- (S4.west);
    \draw[flow] (Pi_prev.east) -- (Si_prev.west);
    \draw[flow] (Pi.east) -- (Si.west);

    \node[mix, anchor=west]
        (M12)
        at ($(S1.east)!0.5!(S2.east)+(1.8cm,0)$)
        {Neighboring-Patch\\Mixer $(P_1,P_2)$};

    \node[mix, anchor=west]
        (M23)
        at ($(S2.east)!0.5!(S3.east)+(1.8cm,0)$)
        {Neighboring-Patch\\Mixer $(P_2,P_3)$};

    \node[mix, anchor=west]
        (M34)
        at ($(S3.east)!0.5!(S4.east)+(1.8cm,0)$)
        {Neighboring-Patch\\Mixer $(P_3,P_4)$};

    \node[dots]
        (D3)
        at ($(S4.east)!0.5!(Si_prev.east)+(3.0cm,0)$)
        {$\vdots$};

    \node[mix, anchor=west]
        (Mi)
        at ($(Si_prev.east)!0.5!(Si.east)+(1.8cm,0)$)
        {Neighboring-Patch\\Mixer $(P_{i-1},P_i)$};

    \node[group_label] at (M12.north)
        {First-Level Interaction};

    \draw[flow] (S1.east) -- (M12.west);
    \draw[flow] (S2.east) -- (M12.west);

    \draw[flow] (S2.east) -- (M23.west);
    \draw[flow] (S3.east) -- (M23.west);

    \draw[flow] (S3.east) -- (M34.west);
    \draw[flow] (S4.east) -- (M34.west);

    \draw[flow] (Si_prev.east) -- (Mi.west);
    \draw[flow] (Si.east) -- (Mi.west);

    \node[dissolved, anchor=west]
        (P1_new)
        at ($(M12.east)+(1.5cm,0.5cm)$)
        {$P'_1$};

    \node[dissolved, below=0.4cm of P1_new]
        (P2_new)
        {$P'_2$};

    \node[dissolved, below=0.4cm of P2_new]
        (P3_new)
        {$P'_3$};

    \node[dissolved, below=0.4cm of P3_new]
        (P4_new)
        {$P'_4$};

    \node[dots, below=0.5cm of P4_new]
        (D4)
        {$\vdots$};

    \node[dissolved, anchor=west]
        (Pi_prev_new)
        at ($(Mi.east)+(1.5cm,0.5cm)$)
        {$P'_{i-1}$};

    \node[dissolved, below=0.4cm of Pi_prev_new]
        (Pi_new)
        {$P'_i$};

    \node[group_label] at (P1_new.north)
        {Patch Update};

    \draw[flow] (M12.east) -- (P1_new.west);
    \draw[flow] (M12.east) -- (P2_new.west);

    \draw[flow] (M23.east) -- (P2_new.west);
    \draw[flow] (M23.east) -- (P3_new.west);

    \draw[flow] (M34.east) -- (P3_new.west);
    \draw[flow] (M34.east) -- (P4_new.west);

    \draw[flow] (Mi.east) -- (Pi_prev_new.west);
    \draw[flow] (Mi.east) -- (Pi_new.west);

    \node[block, anchor=west]
        (BlockA)
        at ($(P1_new.east)!0.5!(P2_new.east)+(1.7cm,0)$)
        {Group $G^{(1)}_1$\\$[P'_1,P'_2]$};

    \node[block, anchor=west]
        (BlockB)
        at ($(P3_new.east)!0.5!(P4_new.east)+(1.7cm,0)$)
        {Group $G^{(1)}_2$\\$[P'_3,P'_4]$};

    \node[block, anchor=west]
        (BlockK)
        at ($(Pi_prev_new.east)!0.5!(Pi_new.east)+(1.7cm,0)$)
        {Group $G^{(1)}_K$\\$[P'_{i-1},P'_i]$};

    \node[dots]
        (D5)
        at ($(BlockB)!0.5!(BlockK)$)
        {$\vdots$};

    \node[group_label] at (BlockA.north)
        {Regrouping};

    \draw[flow] (P1_new.east) -- (BlockA.west);
    \draw[flow] (P2_new.east) -- (BlockA.west);

    \draw[flow] (P3_new.east) -- (BlockB.west);
    \draw[flow] (P4_new.east) -- (BlockB.west);

    \draw[flow] (Pi_prev_new.east) -- (BlockK.west);
    \draw[flow] (Pi_new.east) -- (BlockK.west);

    \node[mix, anchor=west]
        (MixL2)
        at ($(BlockA.east)!0.5!(BlockB.east)+(2.0cm,0)$)
        {Neighboring-Group\\Mixer $(G^{(1)}_1,G^{(1)}_2)$};

    \node[mix, anchor=west]
        (MixK)
        at ($(BlockK.east)+(2.0cm,0)$)
        {Neighboring-Group\\Mixer};

    \node[group_label] at (MixL2.north)
        {Higher-Level Interaction};

    \draw[flow] (BlockA.east) -- (MixL2.west);
    \draw[flow] (BlockB.east) -- (MixL2.west);
    \draw[flow] (BlockK.east) -- (MixK.west);

    \node[mega, anchor=west]
        (MegaL)
        at ($(MixL2.east)+(1.8cm,0.5cm)$)
        {Higher-Level\\Group $G^{(2)}_1$};

    \node[mega, anchor=west]
        (MegaR)
        at ($(MixK.east)+(1.8cm,0)$)
        {Higher-Level\\Group $G^{(2)}_K$};

    \node[group_label] at (MegaL.north)
        {Higher-Level Representations};

    \draw[flow dashed] (MixL2.east) -- (MegaL.west);
    \draw[flow dashed] (MixK.east) -- (MegaR.west);

    \node[mix, anchor=west]
        (FinalMix)
        at ($(MegaL.east)+(1.8cm,-1.5cm)$)
        {Global\\Aggregation};

    \node[patch, right=1.4cm of FinalMix]
        (Output)
        {Hierarchical\\Representation};

    \draw[flow]
        (MegaL.east)
        -|
        (FinalMix.north);

    \draw[flow]
        (MegaR.east)
        -|
        (FinalMix.south);

    \draw[flow]
        (FinalMix.east)
        --
        (Output.west);

    \end{tikzpicture}%
    }

    \caption{
    Architecture of the Hierarchical Neighboring Mixer.
    Patch representations are locally encoded and progressively mixed
    and regrouped to produce a hierarchical representation with an
    expanded temporal receptive field.
    }
    \label{fig:neighboringmixerblock}
\end{figure*}

\subsection{Stationary Wavelet Transform (SWT)}

To expose temporal patterns at different resolutions, NPMixer
decomposes the input using the standard Stationary Wavelet
Transform (SWT)~\cite{dwt,swt}. Unlike the Discrete Wavelet
Transform~\cite{dwt}, SWT does not downsample the filtered signals,
so every coefficient preserves the original sequence length and
remains temporally aligned with the input.

Given $\mathbf{X}\in\mathbb{R}^{C\times L}$, SWT starts from
$\mathbf{A}_0=\mathbf{X}$ and recursively applies fixed low-pass
and high-pass filters. At level $j$, the approximation
$\mathbf{A}_j$ and detail coefficient $\mathbf{D}_j$ are obtained
from $\mathbf{A}_{j-1}$ using the low-pass filter $\mathbf{h}_0$
and high-pass filter $\mathbf{h}_1$, respectively, with dilation
$2^{j-1}$. For decomposition depth $J$, the resulting scale set is
$\mathcal{C}=\{\mathbf{A}_J,\mathbf{D}_J,\ldots,\mathbf{D}_1\}$,
which we equivalently denote as
$\mathcal{C}=\{\mathbf{C}_m\}_{m=0}^{J}$.

The filters are obtained from a predefined wavelet basis and remain
fixed throughout training. SWT therefore provides a stable
multi-resolution decomposition. NPMixer does not apply inverse SWT;
instead, each coefficient is processed independently, and the
resulting representations are combined through learnable scale
fusion.
  
  \subsection{Scale-wise Patching}
  
  For each coefficient
  $\mathbf{C}_m\in\mathbb{R}^{C\times L}$, NPMixer applies the same
  scale-wise processing structure. The coefficient is first divided
  into non-overlapping patches of length $P$, after which an
  intra-patch MLP models local temporal patterns:
  \begin{equation}
  \begin{aligned}
      \mathbf{P}_m
      &=
      \operatorname{Patch}(\mathbf{C}_m)
      \in\mathbb{R}^{C\times N\times P}, \\
      \mathbf{Z}_m
      &=
      \mathbf{P}_m
      +
      \phi_{\mathrm{intra}}^{(m)}(\mathbf{P}_m).
  \end{aligned}
  \end{equation}
  Here, $N$ is the number of patches, and zero-padding is applied
  when $L$ is not divisible by $P$.
  The function $\phi_{\mathrm{intra}}^{(m)}$ is a two-layer MLP
  with GELU activation and dropout, applied independently to each
  patch. Its residual connection preserves the original patch
  representation while learning local temporal structure.
  
  \subsection{Top-$k$ Channel Mixer}
  
  After intra-patch processing, NPMixer applies the Top-$k$ Channel
  Mixer illustrated in Figure~\ref{fig:topk_channel_mixer}.
  Dense channel-dependent modeling may introduce irrelevant
  cross-variate information by mixing every variable
  indiscriminately~\cite{channelmixing}, whereas sparse graph
  construction can effectively uncover hidden dependencies among
  time-series variables~\cite{mtgnn}. Our mixer therefore learns a
  sparse channel relation graph and allows each variable to
  aggregate information only from its most relevant neighbors.
  
  Let
  $\mathbf{Z}_m\in\mathbb{R}^{B\times C\times N\times P}$ denote
  the patch representation of scale $m$. The mixer learns channel
  query and key embeddings
  $\mathbf{Q},\mathbf{K}\in\mathbb{R}^{C\times d_g}$ and computes
  the similarity matrix together with the selected neighborhood:
  \begin{equation}
  \begin{aligned}
      \mathbf{S}
      &=
      \frac{\mathbf{Q}\mathbf{K}^{\top}}{\sqrt{d_g}},
      \qquad
      \mathbf{S}\in\mathbb{R}^{C\times C}, \\
      \mathcal{N}_k(c)
      &=
      \{c\}
      \cup
      \operatorname{TopK}_{j\neq c}
      \bigl(\mathbf{S}_{c,j}\bigr).
  \end{aligned}
  \end{equation}
  The self-channel is explicitly retained. To distinguish it from
  learned cross-channel similarities, its score is replaced by a
  learnable self-logit $\beta$, i.e.,
  $\tilde{s}_{c,c}=\beta$ and
  $\tilde{s}_{c,j}=\mathbf{S}_{c,j}$ for $j\neq c$.
  
  The selected scores are normalized over each channel's Top-$k$ neighborhood
and used as attention weights to aggregate the corresponding neighboring
channel representations. The resulting aggregate is then combined with the
original channel representation through a gated residual update.

The learnable scalar gate $g$ is initialized to a small value, so the mixer
starts close to channel-independent processing and gradually learns how much
cross-channel information to inject. The same learned channel graph is shared
across all SWT scales, providing a consistent dependency structure throughout
the multi-scale pipeline.
  
\paragraph{MLP Channel Mixer.}

We also evaluate a dense MLP-based alternative because the benefit of
channel-dependent modeling varies across datasets. Some benchmarks require
little cross-channel information, whereas others benefit from explicit
interaction~\cite{channelindep,cd2}. We therefore treat channel mixing as a
configurable component rather than assuming that dense or sparse dependence
is universally optimal.

For dense interaction, $\mathbf{Z}_m$ is permuted from
$\mathbb{R}^{B\times C\times N\times P}$ to
$\mathbb{R}^{B\times N\times P\times C}$, and a two-layer channel MLP with
GELU and dropout is applied along the final axis. Its residual contribution
is controlled by a learnable gate $g_{\mathrm{mlp}}$ initialized to a small
value, after which the original tensor order is restored to obtain
$\mathbf{U}_m$.

A small gate approximates channel-independent processing, while a larger
gate enables stronger all-to-all channel interaction, providing a dense
counterpart to the selective Top-$k$ mixer.
  \subsection{Hierarchical Neighboring Mixer Block}

After intra-patch modeling and channel mixing, the
Hierarchical Neighboring Mixer Block captures temporal
dependencies across patches. As illustrated in
Figure~\ref{fig:neighboringmixerblock}, the block first models
relations between neighboring patches and then repeatedly
regroups them into larger temporal blocks. Hierarchical
patch-based models have shown that progressively aggregating
local representations can capture broader temporal
context~\cite{hipatch}.

At hierarchy level $k$, consecutive patches are arranged into
blocks of size $S_k=2^{k-1}P$. For two neighboring blocks
$\mathbf{q}_i^{(k)}$ and $\mathbf{q}_{i+1}^{(k)}$, a
level-specific MLP extracts their relational representation:
\begin{equation}
    \mathbf{R}_i^{(k)}
    =
    \phi_{\mathrm{hier}}^{(k)}
    \left(
        [\mathbf{q}_i^{(k)}
        \parallel
        \mathbf{q}_{i+1}^{(k)}]
    \right),
    \qquad
    S_k=2^{k-1}P,
\end{equation}
where $[\cdot\parallel\cdot]$ denotes concatenation along the
temporal dimension.

The relation is distributed back to both blocks through a
learnable gate $\alpha_k=\sigma(\gamma_k)$:
\begin{equation}
\begin{aligned}
    \widetilde{\mathbf{q}}_i^{(k)}
    &=
    \mathbf{q}_i^{(k)}
    +
    \alpha_k\mathbf{R}_i^{(k)}, \\
    \widetilde{\mathbf{q}}_{i+1}^{(k)}
    &=
    \mathbf{q}_{i+1}^{(k)}
    +
    (1-\alpha_k)\mathbf{R}_i^{(k)}.
\end{aligned}
\end{equation}
The updated representations are regrouped into larger blocks for the next
level, as illustrated in Figure~\ref{fig:neighboringmixerblock}. Repeating
this process progressively expands the receptive field from neighboring
patches to longer temporal contexts. The resulting representation at scale
$m$ is denoted by
$\mathbf{H}_m=\operatorname{HierMix}^{(m)}(\mathbf{U}_m)$, where
$\mathbf{U}_m$ is the output of the channel mixer.

\begin{table*}[!t]
    \caption{Horizon-averaged performance on the seven long-term benchmarks for $H\in\{96,192,336,720\}$. Values for SRSNet, TimeKAN, and Amplifier are taken from SRSNet~\cite{srsnet}; the other baseline values are taken from CycleNet~\cite{cyclenet}. When a source provides standard deviations, only its means are retained for compactness. Boldface identifies the leading value for each metric, and underlining identifies the runner-up. The complete horizon-specific results are reported in Table~C1 of the Supplementary Material.}
    \label{tab:avg-results-standard-datasets-npmixer-srsnet-timekan-amplifier}
    \centering
    \scriptsize
    \setlength{\tabcolsep}{5pt}
    \renewcommand{\arraystretch}{0.95}
    \resizebox{\textwidth}{!}{%
    \begin{tabular}{l|cc|cc|cc|cc|cc|cc|cc|cc|cc}
    \specialrule{1pt}{0pt}{0pt}
    Dataset
    & \multicolumn{2}{c|}{NPMixer}
    & \multicolumn{2}{c|}{SRSNet}
    & \multicolumn{2}{c|}{TimeKAN}
    & \multicolumn{2}{c|}{Amplifier}
    & \multicolumn{2}{c|}{CycleNet}
    & \multicolumn{2}{c|}{PatchTST}
    & \multicolumn{2}{c|}{DLinear}
    & \multicolumn{2}{c|}{Crossformer}
    & \multicolumn{2}{c}{MICN} \\
    \midrule
    & MSE & MAE
    & MSE & MAE
    & MSE & MAE
    & MSE & MAE
    & MSE & MAE
    & MSE & MAE
    & MSE & MAE
    & MSE & MAE
    & MSE & MAE \\
    \midrule

    ETTm1
    & \textbf{0.344} & \textbf{0.376}
    & \underline{0.351} & \underline{0.378}
    & \textbf{0.344} & 0.380
    & 0.353 & 0.379
    & 0.361 & 0.390
    & 0.352 & 0.382
    & 0.357 & 0.379
    & 0.431 & 0.443
    & 0.383 & 0.406 \\

    ETTm2
    & \textbf{0.251} & \textbf{0.313}
    & \underline{0.252} & \underline{0.314}
    & 0.260 & 0.318
    & 0.256 & 0.318
    & 0.270 & 0.324
    & 0.258 & 0.315
    & 0.267 & 0.332
    & 0.633 & 0.578
    & 0.277 & 0.336 \\

    ETTh1
    & \textbf{0.402} & \textbf{0.418}
    & \underline{0.404} & \underline{0.425}
    & 0.409 & 0.427
    & 0.421 & 0.433
    & 0.437 & 0.440
    & 0.417 & 0.430
    & 0.423 & 0.437
    & 0.441 & 0.465
    & 0.433 & 0.462 \\

    ETTh2
    & \textbf{0.309} & \textbf{0.369}
    & 0.334 & 0.386
    & 0.350 & 0.397
    & 0.356 & 0.402
    & 0.371 & 0.407
    & \underline{0.331} & \underline{0.379}
    & 0.431 & 0.447
    & 0.835 & 0.675
    & 0.385 & 0.430 \\

    Electricity
    & \textbf{0.152} & \textbf{0.249}
    & 0.161 & 0.255
    & 0.164 & 0.258
    & 0.163 & 0.256
    & \underline{0.157} & \underline{0.251}
    & 0.162 & 0.254
    & 0.166 & 0.264
    & 0.293 & 0.351
    & 0.181 & 0.291 \\

    Weather
    & \textbf{0.220} & \textbf{0.261}
    & 0.226 & 0.266
    & 0.226 & 0.268
    & \underline{0.224} & \underline{0.264}
    & 0.226 & 0.266
    & 0.230 & 0.265
    & 0.246 & 0.300
    & 0.230 & 0.290
    & 0.245 & 0.298 \\

    Traffic
    & \underline{0.396} & \textbf{0.266}
    & \textbf{0.392} & \underline{0.270}
    & 0.420 & 0.285
    & 0.416 & 0.290
    & 0.413 & 0.281
    & \underline{0.396} & \textbf{0.266}
    & 0.434 & 0.295
    & 0.534 & 0.300
    & 0.535 & 0.312 \\

    \midrule
    \textbf{Win Count}
    & \textbf{6} & \textbf{7}
    & \underline{1} & \underline{0}
    & 1 & 0
    & 0 & 0
    & 0 & 0
    & 0 & 1
    & 0 & 0
    & 0 & 0
    & 0 & 0 \\

    \specialrule{1pt}{0pt}{0pt}
    \end{tabular}%
    }
\end{table*}

\subsection{Scale Projection and Prediction}

The hierarchical patch representation of each scale is flattened
and projected back to the original look-back length. The
projected scales are then combined using normalized learnable
weights:
\begin{equation}
\begin{aligned}
    \widetilde{\mathbf{C}}_m
    &=
    \operatorname{Project}^{(m)}
    \left(
        \operatorname{Flatten}(\mathbf{H}_m)
    \right), \\
    \alpha_m
    &=
    \frac{\exp(\ell_m)}
    {\sum_{r=0}^{J}\exp(\ell_r)}, \qquad
    \mathbf{C}_{\mathrm{fuse}}
    =
    \sum_{m=0}^{J}
    \alpha_m\widetilde{\mathbf{C}}_m .
\end{aligned}
\end{equation}
Here, $\ell_m$ is a learnable logit associated with scale $m$,
and
$\widetilde{\mathbf{C}}_m,
\mathbf{C}_{\mathrm{fuse}}
\in\mathbb{R}^{B\times C\times L}$.

A sequence-level MLP further refines the fused representation through a
residual connection, producing
$\mathbf{Z}
=
\mathbf{C}_{\mathrm{fuse}}
+
\phi_{\mathrm{seq}}(\mathbf{C}_{\mathrm{fuse}})$.
The sequence MLP consists of two linear layers with GELU activation and
dropout.

A final linear prediction layer maps the temporal dimension from the
look-back length $L$ to the forecasting horizon $H$. Inverse RevIN then
restores the original data scale, yielding
$\widehat{\mathbf{Y}}\in\mathbb{R}^{B\times C\times H}$.

\begin{table*}[!t]
    \caption{PEMS performance for each forecasting horizon. Boldface marks the lowest error for a metric, whereas underlining marks the runner-up. Random-seed statistics are reported in Table~C4 of the Supplementary Material.}
    \label{tab:full-results-pems-datasets}
    \centering
    \scriptsize
    \setlength{\tabcolsep}{5pt}
    \renewcommand{\arraystretch}{0.95}
    \begin{tabular}{l|c|cc|cc|cc|cc|cc|cc|cc|cc}
    \specialrule{1pt}{0pt}{0pt}
    Dataset & Horizon
    & \multicolumn{2}{c|}{Ours}
    & \multicolumn{2}{c|}{CycleNet/MLP}
    & \multicolumn{2}{c|}{RLinear}
    & \multicolumn{2}{c|}{iTransformer}
    & \multicolumn{2}{c|}{PatchTST}
    & \multicolumn{2}{c|}{Crossformer}
    & \multicolumn{2}{c|}{DLinear}
    & \multicolumn{2}{c}{SCINet} \\
    \midrule
    &
    & MSE & MAE
    & MSE & MAE
    & MSE & MAE
    & MSE & MAE
    & MSE & MAE
    & MSE & MAE
    & MSE & MAE
    & MSE & MAE \\
    \midrule

    PEMS03
    & 12 & \textbf{0.062} & \textbf{0.166} & \underline{0.066} & \underline{0.172} & 0.126 & 0.236 & 0.071 & 0.174 & 0.099 & 0.216 & 0.090 & 0.203 & 0.122 & 0.243 & \underline{0.066} & \underline{0.172} \\
    & 24 & \textbf{0.083} & \textbf{0.193} & 0.089 & 0.201 & 0.246 & 0.334 & 0.093 & 0.201 & 0.142 & 0.259 & 0.121 & 0.240 & 0.201 & 0.317 & \underline{0.085} & \underline{0.198} \\
    & 48 & 0.133 & 0.247 & 0.136 & 0.247 & 0.551 & 0.529 & \textbf{0.125} & \textbf{0.236} & 0.211 & 0.319 & 0.202 & 0.317 & 0.333 & 0.425 & \underline{0.127} & \underline{0.238} \\
    & 96 & 0.185 & 0.302 & 0.182 & \underline{0.282} & 1.057 & 0.787 & \textbf{0.164} & \textbf{0.275} & 0.269 & 0.370 & 0.262 & 0.367 & 0.457 & 0.515 & \underline{0.178} & 0.287 \\
    \cline{1-18}

    PEMS04
    & 12 & \textbf{0.069} & \textbf{0.171} & 0.078 & 0.186 & 0.138 & 0.252 & 0.078 & 0.183 & 0.105 & 0.224 & 0.098 & 0.218 & 0.148 & 0.272 & \underline{0.073} & \underline{0.177} \\
    & 24 & \textbf{0.083} & \textbf{0.191} & 0.099 & 0.212 & 0.258 & 0.348 & 0.095 & 0.205 & 0.153 & 0.275 & 0.131 & 0.256 & 0.224 & 0.340 & \underline{0.084} & \underline{0.193} \\
    & 48 & \underline{0.106} & \underline{0.221} & 0.133 & 0.248 & 0.572 & 0.544 & 0.120 & 0.233 & 0.229 & 0.339 & 0.205 & 0.326 & 0.355 & 0.437 & \textbf{0.099} & \textbf{0.211} \\
    & 96 & \underline{0.143} & \underline{0.261} & 0.167 & 0.281 & 1.137 & 0.820 & 0.150 & 0.262 & 0.291 & 0.389 & 0.402 & 0.457 & 0.452 & 0.504 & \textbf{0.114} & \textbf{0.227} \\
    \cline{1-18}

    PEMS07
    & 12 & \textbf{0.057} & \textbf{0.156} & \underline{0.062} & \underline{0.162} & 0.118 & 0.235 & 0.067 & 0.165 & 0.095 & 0.207 & 0.094 & 0.200 & 0.115 & 0.242 & 0.068 & 0.171 \\
    & 24 & 0.104 & 0.215 & \textbf{0.086} & \underline{0.192} & 0.242 & 0.341 & \underline{0.088} & \textbf{0.190} & 0.150 & 0.262 & 0.139 & 0.247 & 0.210 & 0.329 & 0.119 & 0.225 \\
    & 48 & \textbf{0.108} & \underline{0.218} & 0.128 & 0.234 & 0.562 & 0.541 & \underline{0.110} & \textbf{0.215} & 0.253 & 0.340 & 0.311 & 0.369 & 0.398 & 0.458 & 0.149 & 0.237 \\
    & 96 & 0.152 & 0.263 & 0.176 & 0.268 & 1.096 & 0.795 & \textbf{0.139} & \underline{0.245} & 0.346 & 0.404 & 0.396 & 0.442 & 0.594 & 0.553 & \underline{0.141} & \textbf{0.234} \\
    \cline{1-18}

    PEMS08
    & 12 & \textbf{0.065} & \textbf{0.169} & 0.082 & 0.185 & 0.133 & 0.247 & \underline{0.079} & \underline{0.182} & 0.168 & 0.232 & 0.165 & 0.214 & 0.154 & 0.276 & 0.087 & 0.184 \\
    & 24 & \textbf{0.085} & \textbf{0.196} & 0.117 & 0.226 & 0.249 & 0.343 & \underline{0.115} & \underline{0.219} & 0.224 & 0.281 & 0.215 & 0.260 & 0.248 & 0.353 & 0.122 & 0.221 \\
    & 48 & \textbf{0.123} & \underline{0.243} & \underline{0.169} & 0.268 & 0.569 & 0.544 & 0.186 & \textbf{0.235} & 0.321 & 0.354 & 0.315 & 0.355 & 0.440 & 0.470 & 0.189 & 0.270 \\
    & 96 & \textbf{0.171} & \underline{0.289} & 0.233 & 0.306 & 1.166 & 0.814 & \underline{0.221} & \textbf{0.267} & 0.408 & 0.417 & 0.377 & 0.397 & 0.674 & 0.565 & 0.236 & 0.300 \\

    \midrule
    \textbf{Win Count}
    &
    & \textbf{10} & \textbf{7}
    & 1 & 0
    & 0 & 0
    & \underline{3} & \underline{6}
    & 0 & 0
    & 0 & 0
    & 0 & 0
    & 2 & 3 \\
    \specialrule{1pt}{0pt}{0pt}
    \end{tabular}
\end{table*}

\section{Experiments}
\begin{table*}[t]
    \centering
    \scriptsize
    \caption{Dataset-wise ablation results, averaged across $H\in\{96,192,336,720\}$. Every model variant is hyperparameter-tuned independently, and the NPMixer row contains the aggregated values used in the main comparison. For each metric, the minimum is boldfaced and the second-lowest value is underlined.}
    \label{tab:ablation_results}
    \resizebox{\textwidth}{!}{
    \setlength{\tabcolsep}{7.0pt}
    \renewcommand{\arraystretch}{1.15}
    \begin{tabular}{l|cc|cc|cc|cc|cc|cc|cc}
    \toprule
    Model
    & \multicolumn{2}{c|}{ETTm1}
    & \multicolumn{2}{c|}{ETTm2}
    & \multicolumn{2}{c|}{ETTh1}
    & \multicolumn{2}{c|}{ETTh2}
    & \multicolumn{2}{c|}{ECL}
    & \multicolumn{2}{c|}{Weather}
    & \multicolumn{2}{c}{Traffic} \\
    \cmidrule{2-15}
    & MSE & MAE
    & MSE & MAE
    & MSE & MAE
    & MSE & MAE
    & MSE & MAE
    & MSE & MAE
    & MSE & MAE \\
    \midrule

    w/o SWT
    & 0.353 & 0.382
    & 0.258 & 0.317
    & 0.418 & 0.429
    & 0.347 & 0.394
    & 0.160 & 0.257
    & 0.223 & 0.264
    & 0.452 & 0.308 \\

    w/o Channel Mixer
    & \underline{0.346} & \textbf{0.375}
    & \underline{0.254} & 0.317
    & 0.415 & 0.427
    & 0.344 & 0.391
    & 0.160 & \underline{0.252}
    & 0.226 & \underline{0.263}
    & \textbf{0.390} & \textbf{0.264} \\

    w/o Neighboring Patch Mixer
    & 0.350 & 0.378
    & \textbf{0.251} & \textbf{0.313}
    & \underline{0.414} & \underline{0.425}
    & \underline{0.337} & 0.386
    & \underline{0.158} & 0.257
    & 0.223 & \textbf{0.261}
    & 0.424 & 0.300 \\

    w/o Intra-Patch Mixer
    & 0.351 & 0.379
    & \underline{0.254} & \underline{0.316}
    & 0.418 & 0.429
    & \underline{0.337} & \underline{0.385}
    & \underline{0.158} & 0.256
    & \underline{0.221} & \textbf{0.261}
    & 0.420 & 0.293 \\

    \midrule

    NPMixer
    & \textbf{0.344} & \underline{0.376}
    & \textbf{0.251} & \textbf{0.313}
    & \textbf{0.402} & \textbf{0.418}
    & \textbf{0.309} & \textbf{0.369}
    & \textbf{0.152} & \textbf{0.249}
    & \textbf{0.220} & \textbf{0.261}
    & \underline{0.396} & \underline{0.266} \\

    \bottomrule
    \end{tabular}
    }
\end{table*}

\subsection{Experimental Setup}

\textbf{Datasets and Baselines.}
The long-term evaluation covers seven benchmarks: ETTh1,
ETTh2, ETTm1, ETTm2, Weather, Electricity, and
Traffic~\cite{informer}. Collectively, these datasets span different
sampling frequencies and numbers of variables. Detailed descriptions
of these benchmarks are provided in Section~A of the Supplementary
Material. We compare NPMixer
with SRSNet~\cite{srsnet}, TimeKAN~\cite{timekan},
Amplifier~\cite{amplifier}, CycleNet~\cite{cyclenet},
PatchTST~\cite{patchtst}, DLinear~\cite{dlinear},
Crossformer~\cite{crossformer}, and MICN~\cite{micn}.

\textbf{NPMixer Variants.}
We evaluate two variants that share the SWT decomposition,
non-overlapping patch processing, intra-patch MLP, hierarchical
neighboring mixer, and learnable scale fusion.
\textbf{NPMixer+TopK} uses the proposed sparse Top-$k$ channel
mixer, whereas \textbf{NPMixer+MLP} uses a dense MLP along the
channel dimension. In the main comparison, NPMixer denotes the
better of the two variants for each dataset and prediction horizon,
selected by MSE; the MAE is taken from the same variant. The selected
variant and hyperparameter configuration for each dataset--horizon
pair are listed in Table~B2 of the Supplementary Material.

\begin{table}[t]
    \centering
    \caption{Top-$k$ sensitivity averaged over prediction lengths
    $H\in\{96,192,336,720\}$. Here, $k=0$
    disables cross-channel mixing. Best and second-best values
    are bold and underlined, respectively.}
    \label{tab:topk-sensitivity}
    \scriptsize
    \setlength{\tabcolsep}{5pt}
    \renewcommand{\arraystretch}{0.85}
    \resizebox{\columnwidth}{!}{%
    \begin{tabular}{ccc|ccc|ccc}
        \toprule
        \multicolumn{3}{c|}{ETTh1}
        & \multicolumn{3}{c|}{Weather}
        & \multicolumn{3}{c}{Traffic} \\
        \cmidrule(lr){1-3}
        \cmidrule(lr){4-6}
        \cmidrule(lr){7-9}
        $k$ & MSE & MAE
        & $k$ & MSE & MAE
        & $k$ & MSE & MAE \\
        \midrule

        0 & 0.438 & 0.439
        & 0 & 0.227 & 0.264
        & 0 & \underline{0.418} & \textbf{0.283} \\

        1 & \textbf{0.432} & \textbf{0.434}
        & 5 & \textbf{0.224} & \underline{0.263}
        & 5 & \textbf{0.414} & \textbf{0.283} \\

        3 & \underline{0.437} & \underline{0.438}
        & 10 & \underline{0.225} & 0.264
        & 20 & 0.432 & 0.298 \\

        5 & 0.438 & 0.440
        & 15 & \textbf{0.224} & \textbf{0.262}
        & 100 & 0.432 & 0.295 \\

        6 & \underline{0.437} & \underline{0.438}
        & 20 & \textbf{0.224} & \underline{0.263}
        & 300 & 0.431 & \underline{0.294} \\

        -- & -- & --
        & -- & -- & --
        & 500 & 0.430 & 0.295 \\

        -- & -- & --
        & -- & -- & --
        & 861 & 0.434 & 0.296 \\

        \bottomrule
    \end{tabular}%
    }
\end{table}

\begin{table}[t]
    \centering
    \caption{
        Average forecasting performance and computational cost of
        non-overlapping (NO) and overlapping (OV) patching with
        $L=336$. MSE/MAE results are averaged over
        $H\in\{96,192,336,720\}$, while parameter counts and epoch
        times are averaged over all horizons. Win Count reports the
        number of best dataset-average MSE/MAE results, with ties
        counted as wins. Best and second-best values are bold and
        underlined, respectively.
    }
    \label{tab:patch-overlap-averaged}

    \scriptsize
    \setlength{\tabcolsep}{4pt}
    \renewcommand{\arraystretch}{1.02}

    \resizebox{0.98\columnwidth}{!}{%
    \begin{tabular}{@{}llcccc@{}}
        \toprule
        Dataset
        & Metric
        & \shortstack{MLP\\NO}
        & \shortstack{MLP\\OV}
        & \shortstack{Top-$k$\\NO}
        & \shortstack{Top-$k$\\OV} \\
        \midrule

        \multirow{3}{*}{ETTh1}
        & Avg. MSE/MAE
        & \textbf{0.4083}/\underline{0.4230}
        & 0.4260/0.4315
        & \textbf{0.4083}/\textbf{0.4228}
        & \underline{0.4258}/0.4315 \\
        & Params (M)
        & \textbf{0.511}
        & \underline{1.479}
        & \textbf{0.511}
        & \underline{1.479} \\
        & Epoch (s)
        & \textbf{4.0}
        & \underline{4.6}
        & 4.9
        & 5.5 \\

        \midrule

        \multirow{3}{*}{Weather}
        & Avg. MSE/MAE
        & \underline{0.2258}/\textbf{0.2640}
        & \textbf{0.2255}/\underline{0.2643}
        & 0.2260/\textbf{0.2640}
        & \underline{0.2258}/0.2645 \\
        & Params (M)
        & \textbf{0.512}
        & \underline{1.480}
        & \textbf{0.512}
        & \underline{1.480} \\
        & Epoch (s)
        & \textbf{78.6}
        & 96.5
        & \underline{93.3}
        & 119.7 \\

        \midrule

        \multirow{3}{*}{Traffic}
        & Avg. MSE/MAE
        & 0.4180/0.2855
        & \textbf{0.4103}/\textbf{0.2783}
        & 0.4178/0.2855
        & \underline{0.4115}/\underline{0.2793} \\
        & Params (M)
        & \textbf{0.570}
        & \underline{1.537}
        & \textbf{0.570}
        & \underline{1.537} \\
        & Epoch (s)
        & \textbf{2654.0}
        & 3321.1
        & \underline{2951.4}
        & 3757.2 \\

        \bottomrule
    \end{tabular}%
    }
\end{table}

\textbf{Evaluation Protocol.}
We evaluate the prediction horizons
$H\in\{96,192,336,720\}$ using the data processing protocol and
train/validation/test splits of TimesNet~\cite{timesnet}.
Following SRSNet~\cite{srsnet}, we consider the same candidate
look-back lengths $L\in\{96,336,512\}$ and select the best
configuration based on validation performance.
This protocol allows each model to use its preferred input length
within the same predefined search space.

Optimization minimizes MSE, whereas forecasting accuracy is
reported using both MSE and MAE. The complete hyperparameter search
spaces are provided in Table~B1 of the Supplementary Material.
For both NPMixer variants, we report the mean over three random
seeds.

\begin{table}[t]
    \centering
    \caption{
        Inference efficiency on ETTh1 with $L=336$, $H=96$,
        and batch size 16. GMACs are measured per sample using
        THOP. Lower values are better.
    }
    \label{tab:npmixer_inference_efficiency}

    \scriptsize
    \setlength{\tabcolsep}{20pt}
    \renewcommand{\arraystretch}{1.08}

    \resizebox{\columnwidth}{!}{%
    \begin{tabular}{@{}lcc@{}}
        \toprule
        Model & Params. (M) & GMACs/sample \\
        \midrule
        SRSNet
        & 0.362
        & 0.006356 \\

        TimeKAN
        & \textbf{0.036}
        & 0.004219 \\

        Amplifier
        & 0.152
        & \textbf{0.000450} \\

        PatchTST
        & 0.082
        & 0.005156 \\

        DLinear
        & \underline{0.065}
        & \underline{0.000456} \\

        Crossformer
        & 4.918
        & 1.025096 \\

        \midrule
        NPMixer+MLP
        & 0.505
        & 0.004181 \\

        NPMixer+Top-$k$
        & 0.505
        & 0.004181 \\
        \bottomrule
    \end{tabular}%
    }
\end{table}

\subsection{Main Results}

Table~\ref{tab:avg-results-standard-datasets-npmixer-srsnet-timekan-amplifier}
reports horizon-averaged performance on the seven long-term
forecasting benchmarks over
$H\in\{96,192,336,720\}$.
NPMixer obtains the lowest average MSE and MAE on ETTm1, ETTm2,
ETTh1, ETTh2, Electricity, and Weather, resulting in the largest
number of metric-wise wins. The gains are particularly noticeable
on ETTh2 and Electricity. On Traffic, NPMixer achieves the lowest
MAE and a tied second-lowest MSE, suggesting that it remains
competitive even for datasets with substantially more variables.

It is also notable that the official SRSNet configurations mostly
use a look-back length of $L=512$%
\footnote{\url{https://github.com/decisionintelligence/SRSNet}},
whereas NPMixer selects $L=336$ for most long-term settings.
Since both methods consider the same candidate look-back lengths
$L\in\{96,336,512\}$, the results suggest that NPMixer can often
achieve lower forecasting errors with a shorter historical context.

Overall, the results indicate that the combination of fixed SWT
decomposition, scale-specific patching, and hierarchical
neighboring-patch aggregation is effective across datasets with
different sampling frequencies and channel dimensions. The
intra-patch module appears to capture local temporal patterns,
while neighboring-patch aggregation may extend the receptive field
without requiring overlapping windows.

Table~\ref{tab:full-results-pems-datasets} reports the short-term
PEMS results. Across the 16 dataset-horizon combinations formed by
PEMS03, PEMS04, PEMS07, and PEMS08, NPMixer ranks first in ten
cases by MSE and seven cases by MAE, with particularly strong
performance at horizons 12 and 24. Although iTransformer and
SCINet remain competitive at several longer horizons, these results
suggest that NPMixer also generalizes well to high-frequency,
short-horizon traffic forecasting.

\subsection{Ablation Study}

Table~\ref{tab:ablation_results} reports the ablation results
averaged over $H\in\{96,192,336,720\}$. The complete model performs
best on most datasets, while removing SWT or the neighboring patch
mixer generally increases errors on ETTm1, ETTh1, ETTh2, and
Traffic. This suggests that aligned multi-scale decomposition and
progressive neighboring-patch aggregation both contribute to
capturing temporal patterns at different ranges. The largest
degradation occurs on ETTh2 when SWT is removed, increasing average
MSE from $0.309$ to $0.347$. The intra-patch mixer is also beneficial
on most datasets, although its effect is smaller on Weather and
Traffic.

The effect of channel mixing is more dataset-dependent. The
\emph{w/o Channel Mixer} variant remains competitive and achieves
the best Traffic results and ETTm1 MAE, consistent with findings
that explicit channel interaction is not always beneficial
\cite{channelindep}. The comparison between sparse Top-$k$ and dense
MLP mixing further suggests that different datasets favor different
cross-channel dependency structures, supporting the use of a
configurable rather than fixed channel mixer.

\subsection{Inference Efficiency}
\label{sec:inference-efficiency}

Table~\ref{tab:npmixer_inference_efficiency} shows that both NPMixer
variants maintain a modest inference cost of 0.004181 GMACs per sample.
Although NPMixer contains more parameters than SRSNet, the second-best
baseline in overall forecasting performance, it requires fewer arithmetic
operations during inference. NPMixer also uses fewer GMACs than TimeKAN,
PatchTST, and Crossformer, while remaining more computationally demanding
than the particularly lightweight Amplifier and DLinear. Overall, its GMAC
requirement remains low, indicating that the additional model capacity does
not result in excessive inference computation. Together with its strong
forecasting performance, these results demonstrate a favorable
accuracy--efficiency trade-off. Nevertheless, GMACs measure arithmetic
operations rather than hardware- and implementation-dependent runtime.

\subsection{Top-$k$ Sensitivity}

Table~\ref{tab:topk-sensitivity} shows that the preferred level of
channel interaction is dataset-dependent. ETTh1 performs best with
$k=1$, Weather benefits from a broader neighborhood with $k=15$,
and Traffic favors sparse or no interaction, with $k=5$ giving the
best average MSE and $k=0$ a tied-best MAE. Larger $k$ values
consistently degrade Traffic performance.

These results support using a controllable Top-$k$ neighborhood
rather than fixed dense channel interaction, since different
datasets require different degrees of cross-variate dependence.

\subsection{Patch Type Analysis}

Table~\ref{tab:patch-overlap-averaged} compares overlapping and
non-overlapping patching using results averaged across all prediction
horizons. Non-overlapping patching performs best on ETTh1 and remains
competitive on Weather, whereas overlapping patching achieves the lowest
average errors on Traffic and the best average MSE on Weather. The win
counts therefore show no consistently dominant patch type across datasets
and metrics.

The computational difference is more pronounced. Overlapping patching
requires approximately $2.7$--$2.9\times$ more parameters and increases
the average per-epoch training time by about $12$--$28\%$. Thus, although
overlapping patches provide accuracy gains on some datasets,
non-overlapping patches maintain competitive average performance with
substantially lower model size and training cost, offering a stronger
overall accuracy--efficiency trade-off.





\section{Conclusion and Future Work}

We proposed NPMixer, a multi-scale forecasting model that combines fixed SWT decomposition, non-overlapping patching, intra-patch modeling, configurable channel mixing, and hierarchical neighboring-patch aggregation. Results on long-horizon benchmarks and short-horizon PEMS tasks demonstrate its effectiveness across datasets with different sampling frequencies and channel dimensions, while ablations confirm the benefits of multi-scale decomposition, neighboring-patch mixing, and dataset-dependent sparse or dense channel interaction. NPMixer nevertheless relies on predefined wavelet, patch, and channel-mixing configurations, remains slower than lightweight linear models, and requires dataset-specific mixer selection. Future work will explore adaptive scale, patch, and channel interaction, more efficient neighboring aggregation, and extensions to probabilistic, irregular, and streaming forecasting.



\bibliography{aaai2027}

@inproceedings{vit,
title={An Image is Worth 16x16 Words: Transformers for Image Recognition at Scale},
author={Alexey Dosovitskiy and Lucas Beyer and Alexander Kolesnikov and Dirk Weissenborn and Xiaohua Zhai and Thomas Unterthiner and Mostafa Dehghani and Matthias Minderer and Georg Heigold and Sylvain Gelly and Jakob Uszkoreit and Neil Houlsby},
booktitle={International Conference on Learning Representations},
year={2021}
}

@inproceedings{patchtst,
title={A Time Series is Worth 64 Words:  Long-term Forecasting with Transformers},
author={Yuqi Nie and Nam H Nguyen and Phanwadee Sinthong and Jayant Kalagnanam},
booktitle={International Conference on Learning Representations},
year={2023}
}

@misc{weather,
  author = {{Max Planck Institute for Biogeochemistry}},
  title = {Jena Climate Dataset},
  year = {2024},
  howpublished = {\url{https://www.bgc-jena.mpg.de/wetter/}},
  note = {Accessed: 2024-05-20}
}

@misc{electricity,
  author       = {Trindade, Artur},
  title        = {{ElectricityLoadDiagrams20112014}},
  year         = {2015},
  howpublished = {UCI Machine Learning Repository},
  note         = {{DOI}: https://doi.org/10.24432/C58C86}
}

@misc{pems,
  author = {{California Department of Transportation (Caltrans)}},
  title = {Performance Measurement System ({PeMS})},
  year = {2024},
  howpublished = {\url{https://pems.dot.ca.gov/}},
  note = {Accessed: 2024-05-20}
}

@book{arima,
  title={Time series analysis: forecasting and control},
  author={Box, George EP and Jenkins, Gwilym M and Reinsel, Gregory C and Ljung, Greta M},
  year={2015},
  publisher={John Wiley \& Sons}
}

@inproceedings{informer,
  title={Informer: Beyond efficient transformer for long sequence time-series forecasting},
  author={Zhou, Haoyi and Zhang, Shanghang and Peng, Jieqi and Zhang, Shuai and Li, Jianxin and Xiong, Hui and Zhang, Wancai},
  booktitle={Proceedings of the AAAI Conference on Artificial Intelligence},
  year={2021}
}

@inproceedings{fedformer,
  title={Fedformer: Frequency enhanced decomposed transformer for long-term series forecasting},
  author={Zhou, Tian and Ma, Ziqing and Wen, Qingsong and Wang, Xue and Sun, Liang and Jin, Rong},
  booktitle={International Conference on Machine Learning},
  year={2022},
  organization={PMLR}
}

@inproceedings{crossformer,
title={Crossformer: Transformer Utilizing Cross-Dimension Dependency for Multivariate Time Series Forecasting},
author={Yunhao Zhang and Junchi Yan},
booktitle={International Conference on Learning Representations},
year={2023}
}

@inproceedings{dlinear,
  title={Are transformers effective for time series forecasting?},
  author={Zeng, Ailing and Chen, Muxi and Zhang, Lei and Xu, Qiang},
  booktitle={Proceedings of the AAAI Conference on Artificial Intelligence},
  year={2023}
}

@article{tsmixer,
title={{TSM}ixer: An All-{MLP} Architecture for Time Series Forecasting},
author={Si-An Chen and Chun-Liang Li and Sercan O Arik and Nathanael Christian Yoder and Tomas Pfister},
journal={Transactions on Machine Learning Research},
year={2023}
}

@article{frets,
  title={Frequency-domain MLPs are more effective learners in time series forecasting},
  author={Yi, Kun and Zhang, Qi and Fan, Wei and Wang, Shoujin and Wang, Pengyang and He, Hui and An, Ning and Lian, Defu and Cao, Longbing and Niu, Zhendong},
  journal={Advances in Neural Information Processing Systems},
  year={2023}
}

@article{waveform, 
title={WaveForM: Graph Enhanced Wavelet Learning for Long Sequence Forecasting of Multivariate Time Series}, 
abstractNote={Multivariate time series (MTS) analysis and forecasting are crucial in many real-world applications, such as smart traffic management and weather forecasting. However, most existing work either focuses on short sequence forecasting or makes predictions predominantly with time domain features, which is not effective at removing noises with irregular frequencies in MTS. Therefore, we propose WaveForM, an end-to-end graph enhanced Wavelet learning framework for long sequence FORecasting of MTS. WaveForM first utilizes Discrete Wavelet Transform (DWT) to represent MTS in the wavelet domain, which captures both frequency and time domain features with a sound theoretical basis. To enable the effective learning in the wavelet domain, we further propose a graph constructor, which learns a global graph to represent the relationships between MTS variables, and graph-enhanced prediction modules, which utilize dilated convolution and graph convolution to capture the correlations between time series and predict the wavelet coefficients at different levels. Extensive experiments on five real-world forecasting datasets show that our model can achieve considerable performance improvement over different prediction lengths against the most competitive baseline of each dataset.}, 
journal={Proceedings of the AAAI Conference on Artificial Intelligence}, 
author={Yang, Fuhao and Li, Xin and Wang, Min and Zang, Hongyu and Pang, Wei and Wang, Mingzhong}, 
year={2023}, 
month={Jun.}}

@inproceedings{msgnet,
  title={Msgnet: Learning multi-scale inter-series correlations for multivariate time series forecasting},
  author={Cai, Wanlin and Liang, Yuxuan and Liu, Xianggen and Feng, Jianshuai and Wu, Yuankai},
  booktitle={Proceedings of the AAAI Conference on Artificial Intelligence},
  year={2024}
}

@inproceedings{
simpletm,
title={Simple{TM}: A Simple Baseline for Multivariate Time Series Forecasting},
author={Hui Chen and Viet Luong and Lopamudra Mukherjee and Vikas Singh},
booktitle={International Conference on Learning Representations},
year={2025}
}

@Inbook{swt,
author="Nason, G. P.
and Silverman, B. W.",
editor="Antoniadis, Anestis
and Oppenheim, Georges",
title="The Stationary Wavelet Transform and some Statistical Applications",
bookTitle="Wavelets and Statistics",
year="1995",
publisher="Springer New York",
address="New York, NY",
pages="281--299",
isbn="978-1-4612-2544-7"
}

@article{dwt,
  title={Continuous and discrete wavelet transforms},
  author={Heil, Christopher E and Walnut, David F},
  journal={SIAM Review},
  volume={31},
  number={4},
  pages={628--666},
  year={1989},
  publisher={SIAM}
}

@inproceedings{timemixer,
	title={TimeMixer: Decomposable Multiscale Mixing for Time Series Forecasting},
	author={Wang, Shiyu and Wu, Haixu and Shi, Xiaoming and Hu, Tengge and Luo, Huakun and Ma, Lintao and Zhang, James Y and ZHOU, JUN},
	booktitle={International Conference on Learning Representations},
	year={2024}
}

@inproceedings{
itransformer,
title={iTransformer: Inverted Transformers Are Effective for Time Series Forecasting},
author={Yong Liu and Tengge Hu and Haoran Zhang and Haixu Wu and Shiyu Wang and Lintao Ma and Mingsheng Long},
booktitle={International Conference on Learning Representations},
year={2024}
}

@inproceedings{micn,
  title={Micn: Multi-scale local and global context modeling for long-term series forecasting},
  author={Wang, Huiqiang and Peng, Jian and Huang, Feihu and Wang, Jince and Chen, Junhui and Xiao, Yifei},
  booktitle={The Eleventh International Conference on Learning Representations},
  year={2023}
}

@inproceedings{tsmixer2,
  title={Tsmixer: Lightweight mlp-mixer model for multivariate time series forecasting},
  author={Ekambaram, Vijay and Jati, Arindam and Nguyen, Nam and Sinthong, Phanwadee and Kalagnanam, Jayant},
  booktitle={Proceedings of the 29th ACM SIGKDD Conference on Knowledge Discovery and Data Mining},
  year={2023}
}

@inproceedings{swintransformer,
  title={Swin transformer: Hierarchical vision transformer using shifted windows},
  author={Liu, Ze and Lin, Yutong and Cao, Yue and Hu, Han and Wei, Yixuan and Zhang, Zheng and Lin, Stephen and Guo, Baining},
  booktitle={Proceedings of the IEEE/CVF International Conference on Computer Vision},
  year={2021}
}

@inproceedings{revin,
title={Reversible Instance Normalization for Accurate Time-Series Forecasting against Distribution Shift},
author={Taesung Kim and Jinhee Kim and Yunwon Tae and Cheonbok Park and Jang-Ho Choi and Jaegul Choo},
booktitle={International Conference on Learning Representations},
year={2022}
}

@article{attention,
  title={Attention is all you need},
  author={Vaswani, Ashish and Shazeer, Noam and Parmar, Niki and Uszkoreit, Jakob and Jones, Llion and Gomez, Aidan N and Kaiser, {\L}ukasz and Polosukhin, Illia},
  journal={Advances in Neural Information Processing Systems},
  year={2017}
}

@article{autoformer,
  title={Autoformer: Decomposition transformers with auto-correlation for long-term series forecasting},
  author={Wu, Haixu and Xu, Jiehui and Wang, Jianmin and Long, Mingsheng},
  journal={Advances in Neural Information Processing Systems},
  year={2021}
}

@misc{timesnet,
      title={TimesNet: Temporal 2D-Variation Modeling for General Time Series Analysis}, 
      author={Haixu Wu and Tengge Hu and Yong Liu and Hang Zhou and Jianmin Wang and Mingsheng Long},
      year={2023},
      eprint={2210.02186},
      archivePrefix={arXiv},
      primaryClass={cs.LG}
}

@misc{channelmixing,
      title={From Similarity to Superiority: Channel Clustering for Time Series Forecasting}, 
      author={Jialin Chen and Jan Eric Lenssen and Aosong Feng and Weihua Hu and Matthias Fey and Leandros Tassiulas and Jure Leskovec and Rex Ying},
      year={2024},
      eprint={2404.01340},
      archivePrefix={arXiv},
      primaryClass={cs.LG},
      url={https://arxiv.org/abs/2404.01340}, 
}

@inproceedings{channelindep,
  title={Channel Dependence, Limited Lookback Windows, and the Simplicity of Datasets: How Biased is Time Series Forecasting?},
  author={Abdelmalak, Ibram and Madhusudhanan, Kiran and Choi, Jungmin and Kl{\"o}tergens, Christian and Yalavarthi, Vijaya Krishna and Stubbemann, Maximilian and Schmidt-Thieme, Lars},
  booktitle={Pacific-Asia Conference on Knowledge Discovery and Data Mining},
  pages={585--597},
  year={2026},
  organization={Springer}
}

@inproceedings{hdmixer,
  title={Hdmixer: Hierarchical dependency with extendable patch for multivariate time series forecasting},
  author={Huang, Qihe and Shen, Lei and Zhang, Ruixin and Cheng, Jiahuan and Ding, Shouhong and Zhou, Zhengyang and Wang, Yang},
  booktitle={Proceedings of the AAAI Conference on Artificial Intelligence},
  volume={38},
  number={11},
  pages={12608--12616},
  year={2024}
}

@article{cyclenet,
  title={Cyclenet: Enhancing time series forecasting through modeling periodic patterns},
  author={Lin, Shengsheng and Lin, Weiwei and Hu, Xinyi and Wu, Wentai and Mo, Ruichao and Zhong, Haocheng},
  journal={Advances in Neural Information Processing Systems},
  volume={37},
  pages={106315--106345},
  year={2024}
}

@inproceedings{mtgnn,
  title={Connecting the dots: Multivariate time series forecasting with graph neural networks},
  author={Wu, Zonghan and Pan, Shirui and Long, Guodong and Jiang, Jing and Chang, Xiaojun and Zhang, Chengqi},
  booktitle={Proceedings of the 26th ACM SIGKDD International Conference on Knowledge Discovery \& Data Mining},
  pages={753--763},
  year={2020}
}

@article{srsnet,
  title={Enhancing time series forecasting through selective representation spaces: A patch perspective},
  author={Wu, Xingjian and Qiu, Xiangfei and Cheng, Hanyin and Li, Zhengyu and Hu, Jilin and Guo, Chenjuan and Yang, Bin},
  journal={Advances in Neural Information Processing Systems},
  volume={38},
  pages={23328--23354},
  year={2026}
}

@article{timekan,
  title={TimeKAN: KAN-based frequency decomposition learning architecture for long-term time series forecasting},
  author={Huang, Songtao and Zhao, Zhen and Li, Can and Bai, Lei},
  journal={arXiv preprint arXiv:2502.06910},
  year={2025}
}

@misc{amplifier,
      title={Amplifier: Bringing Attention to Neglected Low-Energy Components in Time Series Forecasting}, 
      author={Jingru Fei and Kun Yi and Wei Fan and Qi Zhang and Zhendong Niu},
      year={2025},
      eprint={2501.17216},
      archivePrefix={arXiv},
      primaryClass={cs.LG},
      url={https://arxiv.org/abs/2501.17216}, 
}

@article{cd2,
  title={Position: There are no champions in long-term time series forecasting},
  author={Brigato, Lorenzo and Morand, Rafael and Str{\o}mmen, Knut and Panagiotou, Maria and Schmidt, Markus and Mougiakakou, Stavroula},
  journal={arXiv preprint arXiv:2502.14045},
  year={2025}
}

@misc{graphwavenet,
      title={Graph WaveNet for Deep Spatial-Temporal Graph Modeling}, 
      author={Zonghan Wu and Shirui Pan and Guodong Long and Jing Jiang and Chengqi Zhang},
      year={2019},
      eprint={1906.00121},
      archivePrefix={arXiv},
      primaryClass={cs.LG},
      url={https://arxiv.org/abs/1906.00121}, 
}

@misc{stemgnn,
      title={Spectral Temporal Graph Neural Network for Multivariate Time-series Forecasting}, 
      author={Defu Cao and Yujing Wang and Juanyong Duan and Ce Zhang and Xia Zhu and Conguri Huang and Yunhai Tong and Bixiong Xu and Jing Bai and Jie Tong and Qi Zhang},
      year={2021},
      eprint={2103.07719},
      archivePrefix={arXiv},
      primaryClass={cs.LG},
      url={https://arxiv.org/abs/2103.07719}, 
}

@inproceedings{
hipatch,
title={Hi-Patch: Hierarchical Patch {GNN} for Irregular Multivariate Time Series},
author={Yicheng Luo and Bowen Zhang and Zhen Liu and Qianli Ma},
booktitle={Forty-second International Conference on Machine Learning},
year={2025},
url={https://openreview.net/forum?id=nBgQ66iEUu}
}

@article{film,
  title={Film: Frequency improved legendre memory model for long-term time series forecasting},
  author={Zhou, Tian and Ma, Ziqing and Wen, Qingsong and Sun, Liang and Yao, Tao and Yin, Wotao and Jin, Rong and others},
  journal={Advances in Neural Information Processing Systems},
  volume={35},
  pages={12677--12690},
  year={2022}
}

@misc{timebridge,
      title={TimeBridge: Non-Stationarity Matters for Long-term Time Series Forecasting}, 
      author={Peiyuan Liu and Beiliang Wu and Yifan Hu and Naiqi Li and Tao Dai and Jigang Bao and Shu-tao Xia},
      year={2025},
      eprint={2410.04442},
      archivePrefix={arXiv},
      primaryClass={cs.LG},
      url={https://arxiv.org/abs/2410.04442}, 
}


\end{document}